\documentclass[conference, 9pt]{IEEEtran}
\let\IEEEmaketitle=\maketitle
\renewcommand{\maketitle}{\begingroup\let\footnote=\thanks \IEEEmaketitle\endgroup}

\usepackage[]{graphicx}

\usepackage[caption=false,font=footnotesize]{subfig} 

\usepackage{multirow}

\usepackage{amsmath}
\usepackage{bm}

\usepackage{cite}

\usepackage{url}

\usepackage[html,dvipsnames,table]{xcolor}

\usepackage[hidelinks,bookmarks=false,pdfusetitle,pdfdisplaydoctitle,pdfauthor={Mahdi Nazemi et al., University of Southern California}]{hyperref}
\hypersetup{colorlinks,%
    citecolor=ForestGreen,%
    filecolor=Orange,%
    linkcolor=blue,%
    urlcolor=BrickRed,
}

\begin{document}
\bstctlcite{IEEEexample:BSTcontrol}
\IEEEoverridecommandlockouts

\title{NullaNet~Tiny: Ultra-low-latency DNN Inference Through Fixed-function Combinational Logic}

\author{
    \IEEEauthorblockN{
        Mahdi Nazemi*, 
        Arash Fayyazi*\thanks{*M.~Nazemi and A.~Fayyazi contributed equally to this work.}, 
        Amirhossein Esmaili, \\
        Atharva Khare, 
        Soheil Nazar Shahsavani, 
        and Massoud Pedram
    }
    \IEEEauthorblockA{
        School of Electrical and Computer Engineering\\
        University of Southern California, Los Angeles, California 90089-0001\\
        Email: \href{mailto:mnazemi@usc.edu}{\{mnazemi, fayyazi, esmailid, aakhare, nazarsha, pedram\}@usc.edu}
    }
}

\maketitle

\section*{Extended Abstract}

%
%
While there is a large body of research on efficient processing of deep neural networks (DNNs) \cite{DBLP:journals/pieee/SzeCYE17,DBLP:conf/eccv/RastegariORF16,DBLP:conf/nips/HubaraCSEB16,DBLP:journals/corr/ZhouNZWWZ16,DBLP:journals/corr/LiL16,DBLP:conf/iclr/ZhouYGXC17,DBLP:conf/iclr/ZhuHMD17,DBLP:conf/iclr/MishraNCM18,DBLP:journals/corr/abs-1805-06085,DBLP:journals/corr/HanPTD15,DBLP:conf/iclr/ZhuG18,DBLP:conf/eccv/ZhangYZTWFW18,DBLP:journals/corr/abs-1907-04840,DBLP:conf/nips/DingDZGHL19,DBLP:journals/corr/HintonVD15,DBLP:conf/iclr/MishraM18,DBLP:conf/iclr/PolinoPA18,DBLP:journals/corr/abs-1801-05787,DBLP:journals/corr/abs-1805-00907,DBLP:conf/osdi/ChenMJZYSCWHCGK18,DBLP:conf/micro/SharmaPMAKSME16,DBLP:journals/tnn/VenierisB19,DBLP:conf/cvpr/GokhaleJDMC14,DBLP:conf/isca/DuFCILLFCT15,DBLP:conf/fpga/ZhangLSGXC15,DBLP:conf/isca/ChenES16,DBLP:journals/corr/abs-1809-04070,DBLP:conf/isca/KimKCYM16,DBLP:conf/asplos/GaoPYHK17,DBLP:conf/isca/ShafieeNMBSHWS16,DBLP:conf/isca/ChiLXZZLWX16}, ultra-low-latency realization of these models for applications with stringent, sub-microsecond latency requirements continues to be an unresolved, challenging problem. 
Field-programmable gate array (FPGA)-based DNN accelerators are gaining traction as a serious contender to replace graphics processing unit/central processing unit-based platforms considering their performance, flexibility, and energy efficiency. 
NullaNet (2018) \cite{DBLP:conf/aspdac/NazemiPP19}, LUTNet (2019) \cite{DBLP:conf/fccm/WangDCC19}, and LogicNets (2020) \cite{DBLP:conf/fpl/UmurogluAFB20} are among accelerators specifically designed to benefit from FPGAs' capabilities. 

This paper presents NullaNet~Tiny, an across-the-stack design and optimization framework for constructing resource and energy-efficient, ultra-low-latency FPGA-based neural network accelerators. 
The key idea is to replace expensive operations required to compute various filter/neuron functions in a DNN with Boolean logic expressions that are mapped to the native look-up tables (LUTs) of the FPGA device (examples of such operations are multiply-and-accumulate and batch normalization). 

\begin{figure}
    \centering
    \includegraphics[width=\columnwidth]{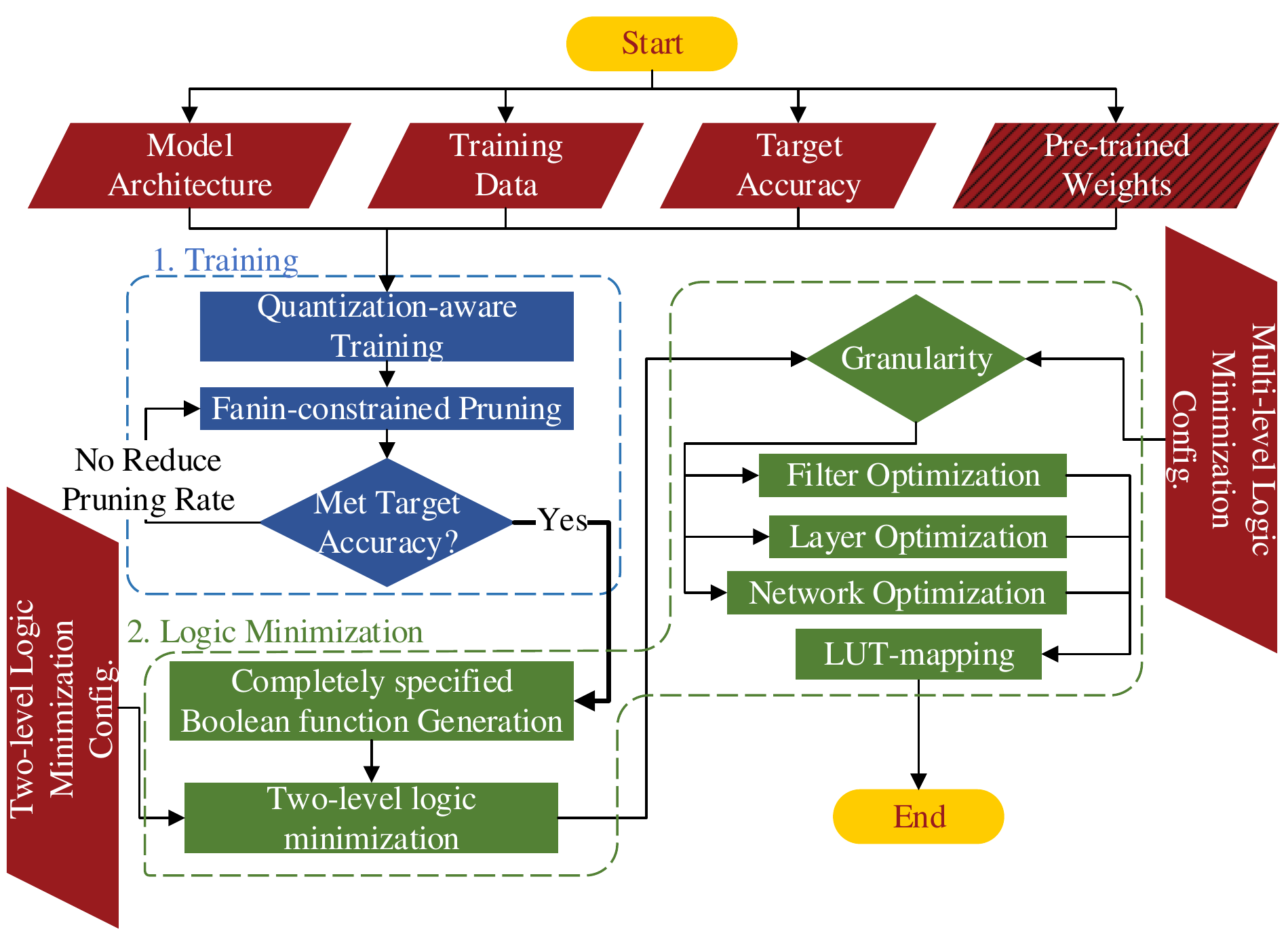}
    \caption{A high-level overview of NullaNet~Tiny's design and optimization flow.}
    \label{fig:nullanet-tiney-flow}
\end{figure}

Fig.~\ref{fig:nullanet-tiney-flow} depicts different steps of NullaNet~Tiny's design and optimization flow. 
The training module is responsible for both quantization-aware training (QAT) and fanin-constrained pruning (FCP) while the logic minimization module performs two-level and multi-level logic minimization in addition to retiming. 
These modules' ultimate goal is to convert each filter/neuron into a highly optimized LUT-based implementation by enumerating all its possible input combinations, recording encountered output values, and optimizing truth tables reflecting those input-output combinations. 

QAT refers to the quantization of activations to binary, bipolar, or multi-bit values during neural network training. 
One of the critical differences between NullaNet~Tiny and prior work is that it employs different activation functions for different layers to yield higher accuracy. 
For example, if the inputs to a DNN assume both negative and positive values, it uses an activation function such as the sign function to capture the range of inputs better. 
On the other hand, if a set of values can only assume non-negative numbers, it relies on the parameterized clipping activation (PACT) \cite{DBLP:journals/corr/abs-1805-06085} function to quantize activations. 
%
%

FCP applies fanin constraints to individual filters/neurons such that the number of inputs to each filter/neuron is small enough to make a realization based on input enumeration (as described in NullaNet \cite{DBLP:conf/aspdac/NazemiPP19}) feasible. 
In this work, FCP is either based on the alternating direction method of multipliers \cite{DBLP:journals/ftml/BoydPCPE11} or gradual pruning \cite{DBLP:conf/iclr/ZhuG18}. 

Finally, functions of different filters/neurons are represented using truth tables which are then fed to the logic minimization module. 
This paper employs the ESPRESSO-II logic minimizer \cite{DBLP:books/sp/BraytonHMS84} for two-level logic minimization and Xilinx Vivado for multi-level logic minimization and retiming. 

Our experimental evaluations summarized in Table~\ref{table:accuracy-hardware-metrics} demonstrate the superior performance of NullaNet~Tiny in terms of accuracy, inference latency, and resource utilization compared to prior art FPGA-based DNN accelerators on the jet substructure classification (JSC) \cite{DBLP:journals/corr/abs-1804-06913} task (the three architectures are based on those described in LogicNets \cite{DBLP:conf/fpl/UmurogluAFB20}). 
\begin{table}
    \caption{Comparison of accuracy \& hardware realization metrics of NullaNet~Tiny and LogicNets \cite{DBLP:conf/fpl/UmurogluAFB20} on the JSC task.}
    \label{table:accuracy-hardware-metrics}
    \centering
    \resizebox{\columnwidth}{!}{%
    \begin{tabular}{l c c c c}
        \hline
        \multirow{2}{*}{\textbf{Arch.}}        & \textbf{Accuracy}  &
        \textbf{LUTs}                          & \textbf{FFs}       &
        \(\bm{f_{max}}\)  \\
        {}                                     & \textbf{(\% Inc.)} &
        \textbf{(Dec.)}                        & \textbf{(Dec.)}    &
        \textbf{(Inc.)}    \\
        \hline
        \multirow{2}{*}{JSC-S}                 & 69.65\%            &
        39                                     & 75                 &
        2,079 MHz          \\
        {}                                     & \textbf{(+1.85)}   &
        \textbf{(5.50\(\bm{\times}\))}         & \textbf{(3.30\(\bm{\times}\))} &\textbf{(1.30\(\bm{\times}\))} \\
        \hline
        \multirow{2}{*}{JSC-M}                 & 72.22\%            &
        1,553                                  & 151                &
        841 MHz            \\
        {}                                     & \textbf{(+1.73)}   &
        \textbf{(9.30\(\bm{\times}\))}         & \textbf{(2.90\(\bm{\times}\))} &\textbf{(1.40\(\bm{\times}\))} \\
        \hline
        \multirow{2}{*}{JSC-L}                 & 73.35\%            &
        11,752                                 & 565                &
        436 MHz            \\
        {}                                     & \textbf{(+1.55)}   &
        \textbf{(3.20\(\bm{\times}\))}         & \textbf{(1.40\(\bm{\times}\))} &\textbf{(1.02\(\bm{\times}\))} \\
        \hline
    \end{tabular}
    }
\end{table}
At about the same level of classification accuracy, compared to Xilinx's LogicNets \cite{DBLP:conf/fpl/UmurogluAFB20}, our design achieves 2.36\(\times\) lower latency and 24.42\(\times\) lower LUT utilization (please note that not all reported clock frequencies are realizable on the target Xilinx VU9P FPGA). 
Similarly, compared to Google's optimized design \cite{DBLP:journals/corr/Coelho20}, our design has a 9.25\(\times\) lower latency. 

\clearpage
\bibliographystyle{IEEEtran}
\bibliography{IEEEabrv,biblio}

\end{document}